\documentclass{bmvc2k}
\newtheorem{proposition}{Proposition}
\newtheorem{theorem}{Theorem}
\newtheorem{corollary}{Corollary}
\usepackage{booktabs} 
\usepackage{multirow}

\usepackage{amsmath}
\usepackage{amssymb}
\usepackage{makecell}

\title{A Super-pixel-based Approach to the Stable Interpretation of Neural Networks}

\addauthor{Shizhan Gong}{szgong22@cse.cuhk.edu.hk}{1}
\addauthor{Jingwei Zhang}{jwzhang22@cse.cuhk.edu.hk}{1}
\addauthor{Qi Dou}{qidou@cuhk.edu.hk}{1}
\addauthor{Farzan Farnia}{farnia@cse.cuhk.edu.hk}{1}

\addinstitution{
 Department of Computer Science and Engineering\\
 The Chinese University of Hong Kong\\
 Hong Kong, China
}

\runninghead{Gong ET AL}{Super-pixel-based Interpretation}

\def\etal{\emph{et al}\bmvaOneDot}

\begin{document}

\maketitle

\begin{abstract}
Saliency maps are widely used in the computer vision community for interpreting neural network classifiers. However, due to the randomness of training samples and optimization algorithms, the resulting saliency maps suffer from a significant level of stochasticity, making it difficult for domain experts to capture the intrinsic factors that influence the neural network's decision. In this work, we propose a novel pixel partitioning strategy to boost the stability and generalizability of gradient-based saliency maps. Through both theoretical analysis and numerical experiments, we demonstrate that the grouping of pixels reduces the variance of the saliency map and improves the generalization behavior of the interpretation method. Furthermore, we propose a sensible grouping strategy based on super-pixels which cluster pixels into groups that align well with the semantic meaning of the images. We perform several numerical experiments on CIFAR-10 and ImageNet. Our empirical results suggest that the super-pixel-based interpretation maps consistently improve the stability and quality over the pixel-based saliency maps. Code is available at: \url{https://github.com/peterant330/SuperPixelGrad}.
\end{abstract}

\section{Introduction}
\begin{figure}[tp]
    \centering
    \includegraphics[width=0.9\linewidth]{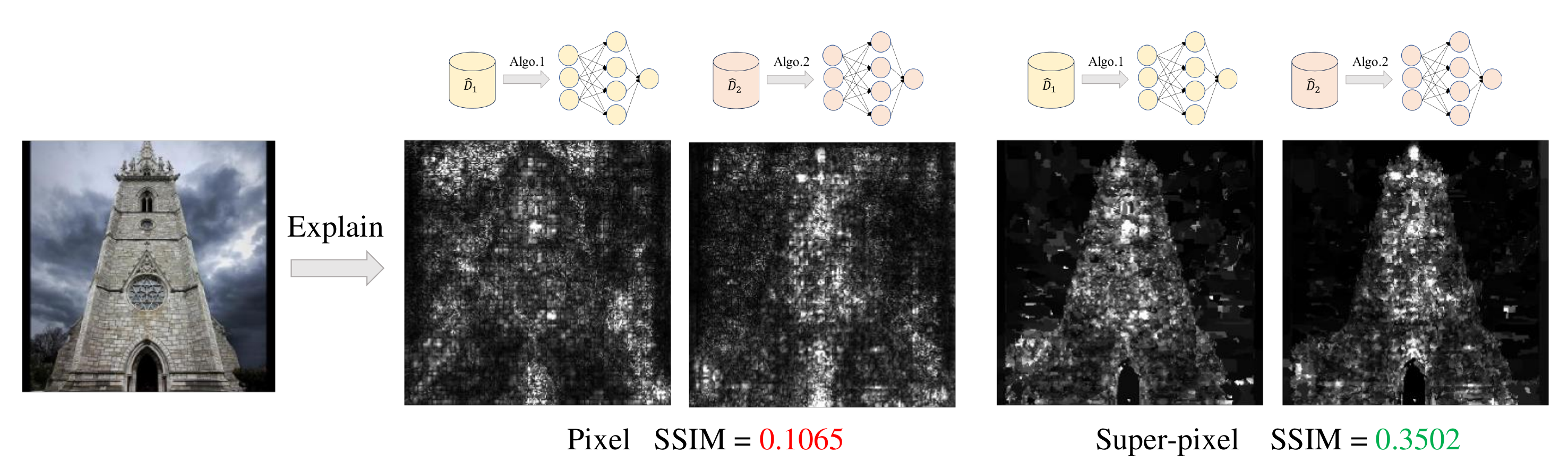}
    \caption{Similarity between saliency maps from two separately trained neural nets: the SSIM between the pixel-based maps is significantly lower than the super-pixel-based maps.}
    \label{fig:teaser}
    \vspace{-3mm}
\end{figure}
Deep learning methods have led to many breakthroughs in computer vision applications, including image classification~\citep{krizhevsky2012imagenet}, object detection~\citep{zhao2019object}, and semantic segmentation~\citep{gong20233dsam}. The complexity of deep neural networks leads to several challenges for interpreting how the models make decisions based on an input, which is of crucial importance in high-stake applications such as automatic driving~\citep{kim2017interpretable} and medical imaging~\citep{gong2023diffusion}. To address this challenge, several explanation methods have been proposed and analyzed within the past few years. A standard approach to explain the predictions of neural networks is to generate \textit{saliency maps} highlighting the image regions with decisive information. The generated saliency maps can also serve as an auxiliary tool for empirically \textit{understanding phenomena} from data-driven models~\citep{freiesleben2022scientific} and further lead to new scientific discoveries and insights. For example, ~\citet{mitani2020detection} demonstrates that saliency maps can accurately highlight the region in fundus images that relate to anaemia. ~\citet{bien2018deep} find that providing deep learning interpretation can improve the performance of clinical experts for knee injury diagnosis from knee MRI.

A popular approach to the generation of interpretation maps is by computing the gradient of the neural net's output prediction to the input~\citep{simonyan2013deep}, which generates saliency maps highlighting the pixels with the strongest local influence on the model's output. However, an important drawback of gradient-based saliency maps is their sensitivity to the randomness in the training process of the neural net classifier, leading to a lack of robustness to the stochasticity of training data and optimization methods. This instability issue can be primarily attributed to the high-dimensional feature spaces of standard image data. As discussed by \cite{smilkov2017smoothgrad}, in the high-dimensional spaces, the calculation of gradients is vulnerable to local fluctuation. Due to the presence of multiple local optima in the loss function of neural networks, stochastic optimization algorithms may converge to different local minima depending on the randomness of the data and algorithm, which may reveal patterns in the interpretation map with noticeable discrepancies. The study by~\cite{woerl2023initialization} reveals the influence of random initialization and stochastic optimization algorithm on the saliency maps and shows different initialization schemes can lead to dissimilar saliency maps. Fig.~\ref{fig:teaser} shows an example of the different pixel-based maps obtained for two neural nets trained on two equally sized and disjoint splits of the ImageNet. This may be less of a problem for applications such as model debugging, where interpretation should reflect the behavior of a specific network. However, it will pose a great threat to using saliency maps for phenomenon understanding~\citep{arun2021assessing}. The significant dissimilarity between the pixel-based maps suggests that the noise in the maps can conceal information about the actual intrinsic factors influencing a phenomenon.

In this paper, we propose a pixel-grouping strategy to boost stability and reduce variance in gradient-based maps, where we partition the pixels into multiple groups to decrease the effective dimension of the input. By taking the average of the saliency map's pixels in every group, we aim to reduce the sensitivity of the resulting map to the stochasticity of the training process. We theoretically show that our proposed grouping strategy for saliency maps will reduce the sensitivity of the generated map to the randomness of training samples, by extending the algorithmic stability framework \citep{bousquet2002stability} to the context of interpretation maps. Following the discussed partitioning strategy, we propose a \emph{super-pixel-based interpretation scheme} by leveraging well-known \textit{super-pixel} schemes in the computer vision literature~\citep{jiao2023transdose,liu2022super,wang2023dynamic}. In a super-pixel scheme, perceptually close pixels sharing homogeneous texture characteristics are clustered together. 
Therefore, we propose to compute the gradient-based maps for the super-pixels of an image, instead of the original pixel-based approach. 

We evaluate the performance of the super-pixel saliency maps by conducting experiments on CIFAR-10 and ImageNet datasets.
Our empirical results suggest that the super-pixel-based extension of gradient-based maps has considerably better stability and generalizability. Also, the numerical results show that the super-pixel-based maps possess relatively higher visual quality and superior interpretability with reasonable tradeoffs with fidelity.

Our main contributions can be summarized as follows:
\begin{itemize}
\item We propose a novel super-pixel-based approach to reduce the variance and noise in saliency maps, which can be flexibly integrated into standard gradient-based interpretation schemes.
\item We theoretically show that the proposed super-pixel-based approach improves the algorithmic stability and reduces the estimation error of the interpretation map.
\item We numerically demonstrate the stability and generalization improvements using our proposed super-pixel-based interpretation maps.
\end{itemize}
\section{Related Work}

\noindent \textbf{Gradient-based Interpretation:} 
Using the gradient of the output of a deep neural network with respect to an input image is a widely used approach to generate saliency maps. This methodology has been applied in a large body of related works with many variants~\citep{smilkov2017smoothgrad,sundararajan2017axiomatic,shrikumar2017learning,selvaraju2017grad,chattopadhay2018generalized,gong2024structured}. Gradient-based saliency maps often suffer from significant noise. Several methods have been developed to reduce the noise and improve the quality of saliency maps. Common techniques include modifying how the gradients pass through the activation function~\citep{zeiler2014visualizing,springenberg2014striving}, Suppressing negative or small activation~\citep{springenberg2014striving,kim2019saliency} and incorporating sparsity prior~\citep{levine2019certifiably,zhang2023moreaugrad}. Nevertheless, these methods do not directly address the noise and fluctuation issues caused by stochastic optimization and limited training data.


\noindent \textbf{Stability Analysis for Interpretation Maps:} 
Many related papers have studied the stability of interpretation methods. For example, ~\citet{arun2021assessing} investigate the repeatability and reproducibility of interpretation methods on medical imaging datasets and find that most of them perform poorly in tests. ~\citet{woerl2023initialization} conduct experiments and observe that neural networks with different initializations generate different saliency maps. \citet{fel2022good} highlight the lack of algorithmic stability in interpretation maps and propose a metric to assess the generalizability of explanation maps. To address this issue, ~\citet{woerl2023initialization} propose a method based on Bayesian marginalization, but it is computationally expensive as it requires training multiple networks. In contrast, our super-pixel-based strategy improves stability by considering randomness from both the training data and process, and it is computationally inexpensive as it only requires a single trained model.


\noindent \textbf{Region-based Interpretation:} Region-based saliency methods, such as RISE~\citep{petsiuk2018rise}, XRAI~\citep{kapishnikov2019xrai}, Score-CAM~\citep{wang2020score}, and Collection-CAM~\citep{ha2022collection}, offer a systematic approach by evaluating the contribution of masks instead of single pixels in predicting the target category. This approach generates an attention map that improves interpretability. However, these methods are computationally intensive, especially when using a large number of masks. In contrast, our approach computes gradients directly with respect to mask indicator variables, similar to gradient-based methods, resulting in similar computation time. Additionally, region-based methods often prioritize visual quality and aggregate features through random or dynamic mask generation, which may lack stability. In contrast, our method utilizes a static grouping based on semantic meanings, leading to significantly improved interpretation stability.
\newtheorem{definition}{Definition}

\section{Methodology}
\subsection{Gradient-based Saliency Maps}
Throughout this work, we denote the input to the neural network with vector $\mathbf{x} \in \mathbb{R}^d$ and the output with vector $\mathbf{y} \in \mathbb{R}^C$. Note that $\mathbf{y}$ is the post-softmax layer and hence contains the assigned likelihood for each of the $C$ classes in the classification task. The neural network is trained to compute a map $f_\theta:\mathbb{R}^d \rightarrow \mathbb{R}^C$ that maps the input vector to the output vector, where $\theta \in \Theta$ is the parameters of the neural network and $\Theta$ is the feasible set of the parameters. 

We use $\theta^* \in \Theta$ to denote the solution under the population distribution of test data. However, in practice, we only have access to a limited set of $n$ training samples, through which we obtain $\hat{\theta}$ minimizing the empirical risk. We note that unlike the deterministic $\theta^*$, $\hat{\theta}$ is indeed a random vector affected by the stochasticity of the training process, such as the random initialization and batch selection in stochastic gradient methods.

\begin{definition}
  The Simple~Gradient (SG) generates a saliency map by taking derivatives of the output of the neural network with respect to the input:
$SG(f_{\hat{\theta},c}, \mathbf{x}):=\nabla_{\mathbf{x}}f_{\hat{\theta},c}(\mathbf{x}),$
where $c$ is chosen to be the neural network’s predicted label with the maximum prediction score.  
\end{definition}

 Given infinitely many training data and an oracle solving the optimization problem with zero error, we would have $SG(f_{\theta^*,c}, \mathbf{x})$, which excludes the influence of limited training samples and stochastic training process. 
To understand the effect of finite training sets on the interpretation, we analyze the generalization of the interpretation maps. We define an interpretation loss that reflects how much the estimated saliency map deviates from the population saliency map, which is the $L_2$-norm of the difference between $SG(f_{\hat{\theta},c}, \mathbf{x})$ and $SG(f_{\theta^*,c}, \mathbf{x})$. The interpretation loss can be extended to any interpretation method $I$ with saliency maps:
\begin{equation}
\mathcal{L}(I, f_{\hat{\theta}}, \mathbf{x}) :=\bigl\Vert I(f_{\hat{\theta}}, \mathbf{x}) - I(f_{\theta^*}, \mathbf{x})\bigr\Vert,
\end{equation}
Therefore, our goal is to reduce the expected norm difference between the empirical $I(f_{\hat{\theta}}, \mathbf{X})$ and population  $I(f_{\theta^*}, \mathbf{X})$ interpretation. 

Note our definition of the interpretation loss based on the assumption that there is a unique population saliency map, which is generally true for classification networks trained with cross-entropy according to the directional convergence theory~\citep{ji2020directional}. For some special cases when the assumption is violated, we can further extend the optimal function $f^*$ to be the expectation of the optimal population solution according to a uniform distribution on the set of optimal solutions $\Theta^*$:  $f^*(x)=\mathbb{E}_{\theta\sim \mathrm{unif}(\Theta^*)}[f_{\theta}(x)]$.

To evaluate the generalizability of the interpretation method to unseen data, we define the interpretation generalization error as the expected gap of interpretation loss between the training and population distribution. The expectation is taken over both the random dataset sampled from the population distribution and the stochasticity of the training algorithm:
\begin{definition}[Interpretation Generalization Error] For an interpretation method $I$ and a random dataset of size $n$, the generalization error of the interpretation is defined as:
\begin{equation*}
\epsilon_{\text{gen}}(I) :=
\Bigl\vert\mathbb{E}_{\hat{\theta}, \mathbf{x}_i}\Bigl[\mathbb{E}_{\mathbf{x}\sim p(\mathbf{x})} \bigl[\mathcal{L}(I, f_{\hat{\theta}}, \mathbf{x})\bigr] \\
\;\;-\frac{1}{n}\sum_{i=1}^n \mathcal{L}(I, f_{\hat{\theta}}, \mathbf{x}_i)\Bigr]\Bigr\vert,
\end{equation*}
\end{definition}
where $p(\mathbf{x})$ is the distribution of $\mathbf{x}$, $n$ is the size of the training data and $\mathbf{x}_i$ denotes the $i$-th training sample.

To follow standard algorithmic stability-based generalization analysis \citep{bousquet2002stability}, we define the uniform stability  of an interpretation algorithm as follows:
\begin{definition}[$\epsilon$-uniformly stable] An interpretation algorithm $I$ is called $\epsilon$-uniformly stable if for every $f_{\hat{\theta}_1}$ and $f_{\hat{\theta}_2}$ trained with dataset $S$, $S'$ such that $S$ and $S'$ differ in at most one sample, we have
\begin{equation}
\sup_\mathbf{x}\mathbb{E}_{S,S',A}\Bigl[\bigl\Vert I(f_{\hat{\theta}_1}, \mathbf{x})-I(f_{\hat{\theta}_2}, \mathbf{x})\bigr\Vert\Bigr]\, \leq\, \epsilon.
\end{equation}
Here, the expectation is taken over the randomness of training sets $S$, $S'$, and the training algorithm $A$.
\end{definition}
We intuitively expect that for two datasets that are different in only one sample, a generalizable interpretation scheme would result in similar outputs for the trained neural networks. Consistent with this intuition, we prove the following theorem connecting the uniform stability and generalization error of an interpretation method: 
\begin{theorem}\label{Theorem gen}
Suppose an interpretation scheme $I$ is $\epsilon$-uniformly stable. Then, the following generalization bound holds:
$\epsilon_{\text{gen}} (I) \, \leq\, \epsilon.$

\label{theorem}
\end{theorem}
We defer the proofs to the Appendix. As Theorem~\ref{Theorem gen} suggests, improving the stability of the interpretation map will boost its generalizability to unseen data.

\subsection{Controlling Generalization Error via Partitioning the Pixels}
To reduce the generalization error, we propose to partition pixels and then perform a gradient-based interpretation. According to our proposal, we calculate the input-based gradient with respect to the pixel group indicators instead of individual pixels. 
Here, we partition the pixels into $p$ groups, denoted as $\mathbb{S}(\mathbf{x}):=\{S_1(\mathbf{x}), \cdots, S_p(\mathbf{x})\}$. In our analysis, we omit $\mathbf{x}$ from the notation for simplicity. We let $\mathbf{g} = \mathbf{0}_p$ to be  the $p$-dimensional zero vector, and define matrix $\mathbf{A} \in \{0, 1\}^{d\times p}$ such that for the $(i,j)$-th entry $a_{i,j}=1$ if the $i$-th pixel belongs to $S_j$ and $a_{i,j}=0$ otherwise. We define the grouped version of Simple Gradient ($g-SG$), where the saliency map is generated by taking gradient with respect to $\mathbf{g}$ and then broadcast to the pixels belonging to the groups:
$g\text{-}SG(f_{\hat{\theta}},\mathbf{x}) := \mathbf{A}^T\nabla_{\mathbf{g}}f_{\hat{\theta}}(\mathbf{x} + \mathbf{A}\mathbf{W}\mathbf{g}),$
where $\mathbf{W} = \text{diag}([\frac{1}{\vert S_1 \vert}, \cdots, \frac{1}{\vert S_p \vert}])$ is a diagonal matrix composed of weights related to the number of pixels within each group. The above expression can be further simplified as
$g\text{-}SG(f_{\hat{\theta}},\mathbf{x}) = \kappa(\mathbf{x}) \nabla_\mathbf{x}f_{\hat{\theta}}(\mathbf{x}),$
where $\kappa(\mathbf{x}) \in \mathbb{R}^{d\times d}$ is defined as:
\begin{equation}
\kappa(\mathbf{x})_{i,j} := \begin{cases}
\frac{1}{\vert S_k\vert}& i,j \in S_k\\
0& \text{otherwise.}
\end{cases}
\end{equation}
Note that the partitioned Simple Gradient calculates the mean value of the importance score within a group, and uses the mean value as the map score for all the group's pixels. This averaging operation will reduce the interpretation loss as stated in the following propositions.
\begin{proposition} The interpretation loss of partitioned Simple Gradient is upper-bounded by that of standard Simple Gradient, i.e.,
$\forall f, \mathbf{x}:\quad \mathcal{L}(g\text{-}SG, f, \mathbf{x}) \leq \mathcal{L}(SG, f, \mathbf{x}).$
\label{prop:1}
\end{proposition}
This proposition is shown by applying Jensen's inequality to the convex norm-squared function. We also prove the following proposition providing the closed-form expression of the gap between the norm-squared-based loss functions.
\begin{proposition} The difference between $\mathcal{L}^2(SG, f_{\hat{\theta}}, \mathbf{x})$ and $\mathcal{L}^2(g\text{-}SG, f_{\hat{\theta}}, \mathbf{x})$ can be simplified as follows where $\mathrm{Var}(\cdot)$ denotes the empirical variance of an input vector:
\begin{equation*}
\Delta\mathcal{L}^2(SG, f_{\hat{\theta}}, \mathbf{x}):=\mathcal{L}^2(SG, f_{\hat{\theta}}, \mathbf{x}) - \mathcal{L}^2(g\text{-}SG, f_{\hat{\theta}}, \mathbf{x}) 
= \sum_{S\in\mathbb{S}}\vert S\vert \cdot \mathrm{Var}(\nabla (f_{\hat{\theta}}-f_{\theta^*})(\mathbf{x})|_S).
\label{equal:gap}
\end{equation*}
\label{prop:2}
\end{proposition}
Our next result indicates the improved stability degree of an interpretation map after applying the proposed partitioning scheme. 
\begin{proposition} If $SG$ for a neural network $f$ is $\epsilon$-uniformly stable, then its grouping version $g\text{-}SG$ is $\epsilon'$-uniformly stable for a certain $\epsilon'\leq\epsilon$.
\label{prop:3}
\end{proposition}
Combining the above result with  Theorem~\ref{theorem}, we can show the following corollary:
\begin{corollary} If $SG$ on a neural network $f$ is $\epsilon$-uniformly stable, then its generalization error can be reduced by the grouping of pixels, i.e.,
$\epsilon_{\text{gen}}(g\text{-}SG) \leq \epsilon_{\text{gen}}(SG).$
\label{prop:4}
\end{corollary}
Note that the proposed grouping strategy can be adapted to other well-known variants of the Simple Gradient. While the above theoretical results are shown for the Simple Gradient, they can be extended to other gradient-based schemes that generate a single saliency map. Moreover, for composite saliency maps such as SmoothGrad, as long as the partition is fixed based on the original image, the same properties still hold due to the linearity of expectation and integration. In practice, flexible and dynamic partition gives even more promising results. We defer detailed analysis regarding SmoothGrad, Integrated Gradients, and Sparsified SmoothGrad to the Appendix. Based on the discussion, we need to find a sensible grouping scheme such that pixels within each assigned group are semantically related. The next section discusses a pixel clustering-based approach to this task.

\subsection{A Super-Pixel-Based Partitioning Strategy}
As we set the importance score within a group to be identical, the pixels of the same group will have the same effect on the output. Therefore, we expect every partition's pixels to be closely located and share the same texture and semantic information. A standard tool in computer vision to achieve these properties is to cluster pixels using the super-pixel algorithms.

Super-pixel refers to the perceptual grouping of pixels. This goal can be obtained by clustering pixels according to pixel characteristics such as brightness, intensity, and color. Several algorithms have been established to generate super-pixels such as Felzenszwalb's method~\citep{felzenszwalb2006efficient}, Quickshift~\citep{vedaldi2008quick}, SLIC~\citep{achanta2012slic}, and Compact watershed~\citep{neubert2014compact}. Felzenszwalb uses a graph-based approach with a minimum spanning tree algorithm. Quickshift employs a mode-seeking algorithm with kernelized mean-shift. SLIC applies K-means in the 5D space of color and location data. Compact Watershed visualizes the image as a topographic surface and uses a flooding process to form regions from local minima. In practice, pixels within a super-pixel have been usually observed to be semantically relevant and often belong to the same object. Therefore, the assumption that pixels inside a super-pixel have the same effect on the model's prediction and interpretation sounds relevant. 

Furthermore, standard super-pixel algorithms perform an unsupervised learning process without using any label information. Therefore, the super-pixel identification process will be statistically independent of the neural network's risk function. Although the intensity and importance score within a super-pixel can be related, the pixel-wise interpretation noise could be nearly independent, suggesting that applying super-pixels for grouping the pixels can significantly reduce the generalization error. We empirically test this hypothesis in the Appendix.

\begin{figure}
\begin{tabular}{cc}
\bmvaHangBox{\parbox{7.4cm}{~\\[2.8mm]
\rule{0pt}{1ex}\hspace{-5mm}\includegraphics[width=0.65\textwidth]{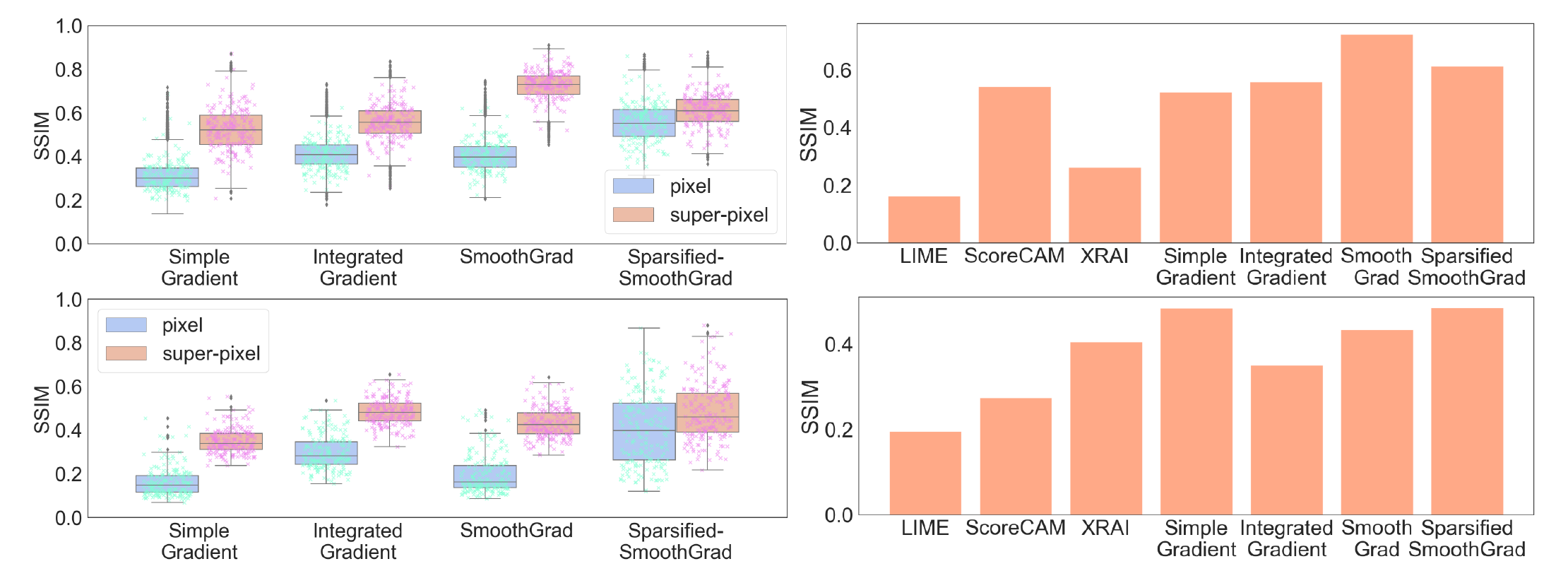}\\[-0.1pt]}}&
\bmvaHangBox{\parbox{2.7cm}{~\\[0mm]
\rule{0pt}{1ex}\hspace{-5mm}\resizebox{0.37\textwidth}{!}{\begin{tabular}{l|cc}
\toprule
\multirow{2}{*}{Methods}&\multicolumn{2}{c}{MeGe (\%)$\uparrow$} \\
\cline{2-3}
&pixel.&super.\\
\midrule
Simple Gradient& 13.28$_{\pm 0.14}$&\textbf{24.98}$_{\pm 0.18}$\\
Integrated-Gradients&15.46$_{\pm 0.19}$&\textbf{33.57}$_{\pm 0.24}$\\
SmoothGrad&17.62$_{\pm 0.04}$&\textbf{48.84}$_{\pm 0.24}$\\
SparsifiedSmoothGrad&17.64$_{\pm 0.04}$&\textbf{22.98}$_{\pm 0.13}$\\
\midrule
LIME&\multicolumn{2}{c}{5.90$_{\pm 0.35}$}\\
ScoreCAM&\multicolumn{2}{c}{15.58$_{\pm 0.52}$}\\
XRAI&\multicolumn{2}{c}{7.30$_{\pm 0.22}$}\\
\bottomrule
\end{tabular}\\[-0.1pt]}}}
\end{tabular}
\caption{\textbf{Left}: The SSIM of saliency map between two
models trained with disjoint training datasets or different initialization from CIFAR10 (top) and Imagenet
(bottom). \textbf{Middle}: Comparison of SSIM with region-based interpretation. \textbf{Right}: Comparison of MeGe between pixel-based and super-pixel-based methods. $L_2$-norm is used as the distance measure.}
\label{fig:ssim}
\end{figure}

\section{Experimental Evaluation}

We performed extensive numerical experiments to evaluate the stability, generalizability, fidelity, interpretability, and visual quality of the proposed super-pixel-based interpretation and their comparison to baselines. We defer more experiments such as inter-architecture reproducibility and cost analysis to the Appendix. 
\subsection{Stability and Generalizability}
\label{cifar}

\begin{figure*}[tp]
    \centering
    \includegraphics[width=0.9\linewidth]{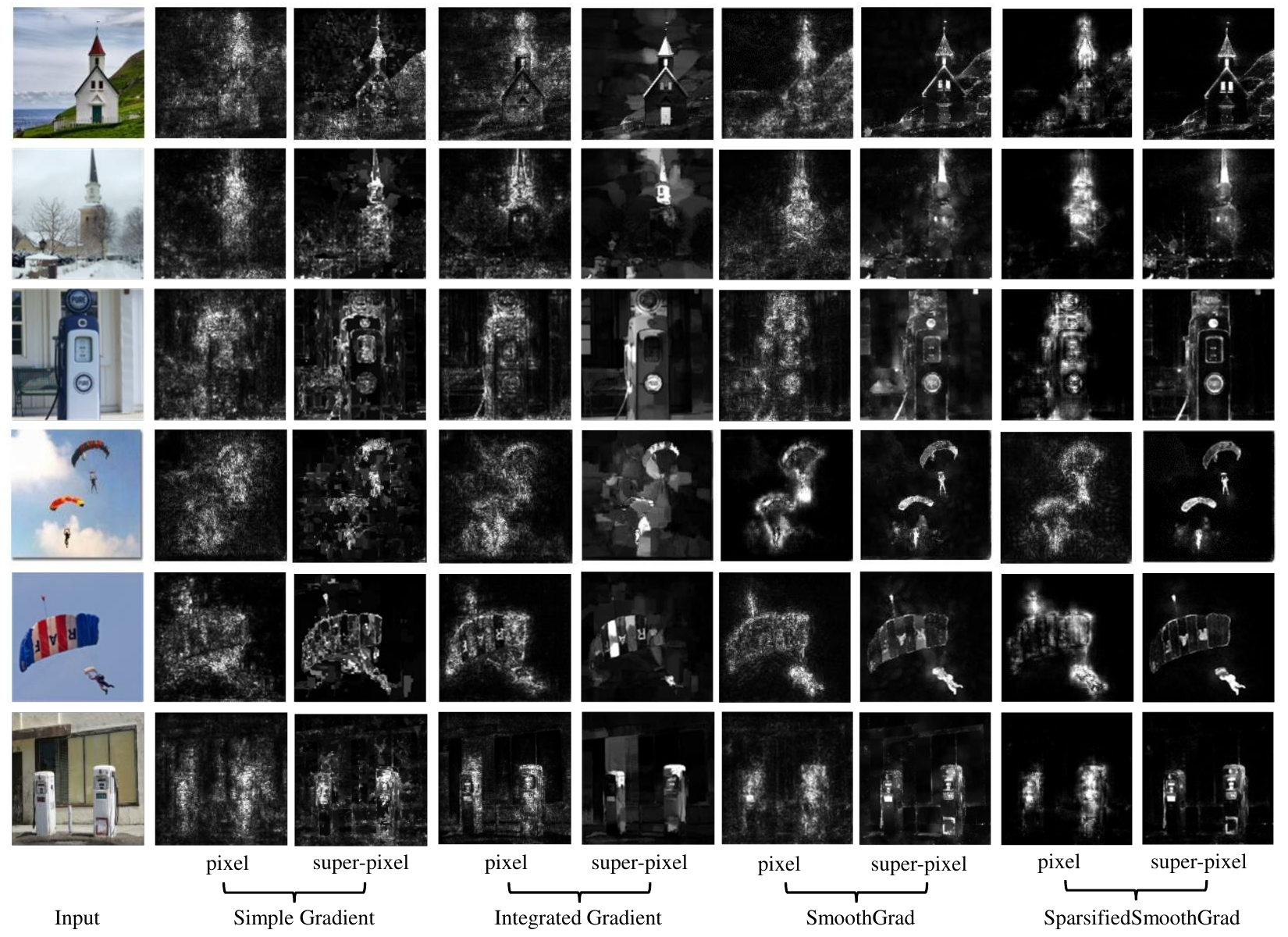}
    \caption{Qualitative comparison on ImageNet dataset between pixel-based and super-pixel-based interpretation maps for different gradient-based methods.}
    \label{fig:vis_imagenet}
\end{figure*}

To validate that the super-pixel-based interpretation map improves the stability properties, we ran numerical experiments on CIFAR-10 and ImageNet. For CIFAR-10, we split the training set into disjoint subsets and train separate Resnet-18~\citep{he2016deep}. For ImageNet, we downloaded two pre-trained Efficientnet-B0~\citep{tan2019efficientnet} from different packages. We then used the two networks to generate saliency maps on test data and used the Structural Similarity Index Measure (SSIM) as the metric for evaluating the similarity between the maps. We used the SLIC algorithm with 100 segments and Quickshift with a maximum distance of 3 for super-pixel generation on CIFAR-10 and ImageNet respectively. The experimental results are shown in Fig.~\ref{fig:ssim}. For almost all the interpretation methods, the SSIM score is relatively low. On the other hand, using super-pixels led to a significant improvement in the SSIM score.
We also evaluated the generalizability of our method using the mean generalizability (MeGe) metric~\cite{fel2022good} on CIFAR-10, which utilizes 5-fold division to measure the change of explanation when a sample is removed from the training set and therefore reflects the generalizability of the interpretation method. We used $L_2$-norm as the distance measure, which aligns with our definition of generalization loss. The findings in Fig.~\ref{fig:ssim} validate our theoretical results that the super-pixel grouping enhances the generalization of interpretation maps.
Additionally, we compared our methods with region-based interpretation methods including LIME, Score-CAM, and XRAI, and found that our methods still exhibit superior stability and generalization, likely due to our static partition based on semantics. Visualizations of gradient-based saliency maps in Fig.~\ref{fig:vis_imagenet} further demonstrate that our super-pixel-based methods generate higher-quality saliency maps, reducing noise and improving visual clarity.

\subsection{Selection of the Super-Pixel Algorithm}
\begin{figure*}[t!]
    \centering
    \includegraphics[width=0.8\linewidth]{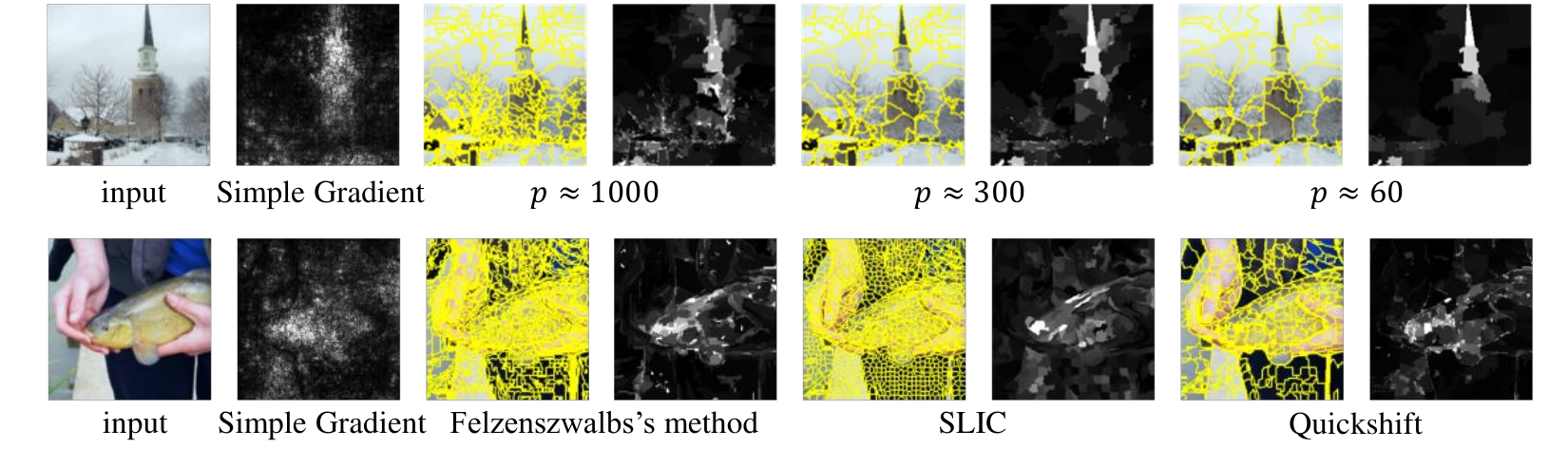}
    \caption{Visualization of super-pixel-based simple gradient maps with different numbers of groups in Quickshift (\textbf{Top}) and different super-pixel algorithms (\textbf{Bottom}).}
    \label{fig:ablation}
\end{figure*}

We analyzed the impact of different super-pixel schemes on the proposed super-pixel-based saliency maps. We varied the super-pixel generating algorithm and size, and performed experiments on Simple Gradient maps with ImageNet. The results, shown in Fig.~\ref{fig:ablation}, indicate that the choice of super-pixel scheme has minimal effect on the method's performance. Although the saliency maps differ slightly, they consistently exhibit higher visual quality than the original Simple Gradient maps. Additionally, we compared saliency maps generated using different super-pixel sizes. Smaller super-pixels highlight fine-grained details, while larger super-pixels produce more compact and sharper saliency maps.

\subsection{Trade-off with Fidelity}
We evaluated the fidelity of the super-pixel-based approach compared to the pixel-based simple-grad maps on ImageNet. We report the following fidelity metrics: deletion~\citep{petsiuk2018rise}, insertion~\citep{petsiuk2018rise}, and $\mu$fidelity~\citep{bhatt2020evaluating}. Deletion measures a decrease in predicted class probability as pixels are removed (most relevant first, MoRF). A low area under the probability curve indicates a good explanation. Insertion is the opposite, which measures the increase in predicted class probability as more pixels are added to the image (MoRF), starting from a blank image. The $\mu$fidelity is the correlation between a drop in predicted probability and the average importance score of the random masked regions. As our method is not designed to improve the fidelity of the interpretation method and a common intuition lies in that there is a trade-off between stability and fidelity, the super-pixel-based method does not significantly drop the fidelity score (Table~\ref{table:fidelity}). We also compare the super-pixel-based partitioning with random partitioning and square-patch partitioning, where we kept the number of groups the same. The results suggest the proposed method achieves the highest fidelity score, suggesting super-pixels align better with the semantics of images and perform better than random partitioning.

 \begin{table*}[tp]

\footnotesize
\centering
\renewcommand\arraystretch{1}
\begin{center}
\resizebox{1\textwidth}{!}{
\begin{tabular}{l|p{5.5mm}p{5.5mm}p{5.5mm}p{5.5mm}p{0.5mm}p{5.5mm}p{5.5mm}p{5.5mm}p{5.5mm}p{0.5mm}p{5.5mm}p{5.5mm}p{5.5mm}p{5.5mm}}
\toprule
\multirow{2}{*}{Methods\rule{0pt}{2.4ex}}&\multicolumn{4}{c}{Deletion(\%)$\downarrow$} && \multicolumn{4}{c}{Insertion(\%)$\uparrow$}&&\multicolumn{4}{c}{$\mu$Fidelity(\%)$\uparrow$} \\
\cline{2-5}
\cline{7-10}
\cline{12-15}
&pixel&sup. &square&rand&&pixel&sup. &square&rand&&pixel&sup. &square&rand\\
\midrule
Simple Gradient&14.0&\textbf{13.8}&21.4&23.0&&54.2&\textbf{62.9}&58.0&42.9&&\textbf{27.9}&27.2&26.8&24.1\\
Integrated Gradient&\textbf{11.3}&13.4&18.5&21.2&&60.5&\textbf{64.4}&61.8&49.3&&30.8&\textbf{31.7}&30.4&4.6\\
SmoothGrad&\textbf{11.1}&\textbf{11.1}&20.9&20.4&&60.3&\textbf{63.7}&60.9&44.8&&\textbf{32.8}&32.2&29.0&28.6\\
SparsifiedSmoothGrad&14.9&\textbf{13.7}&19.3&25.1&&\textbf{61.9}&61.6&60.7&45.9&&\textbf{33.3}&32.1&29.1&31.3\\
\bottomrule
\end{tabular}
}
\end{center}
\caption{Comparison of fidelity among pixel-based, super-pixel (sup.), square patch, and random partition. The metrics are based on 200 random samples from the ImageNet.}
 \label{table:fidelity}
\end{table*}

\begin{figure*}[t!]
    \centering
    \includegraphics[width=\linewidth]{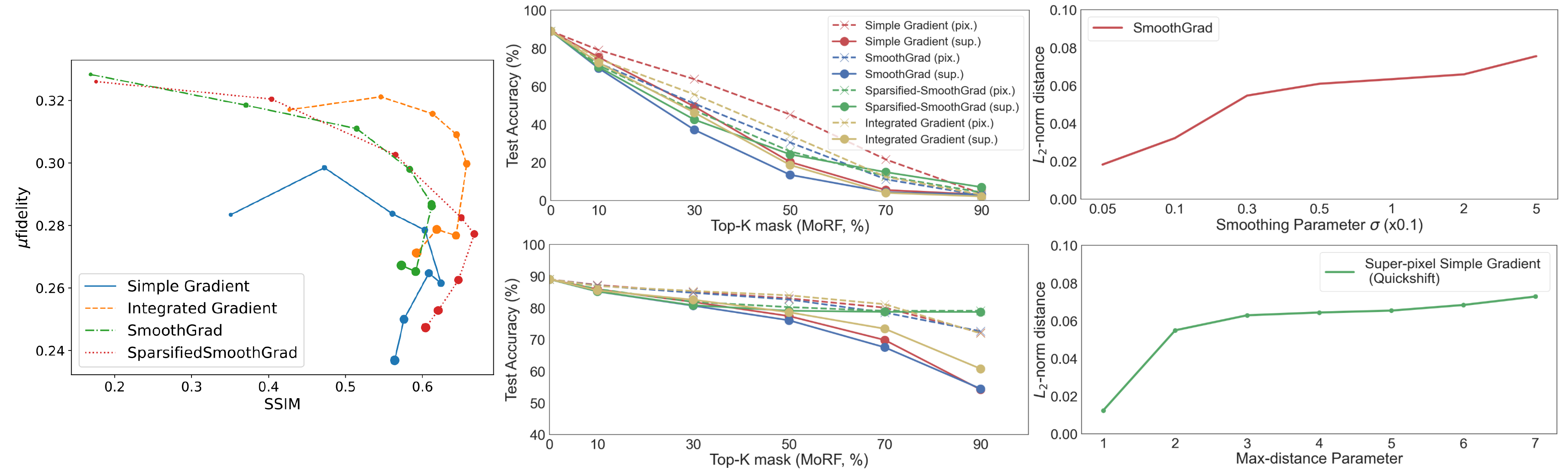}
    \caption{\textbf{Left}: Tradeoff curve between SSIM and $\mu$Fidelity. The size of the dots reflects the size of super-pixels. \textbf{Middle}: Comparison of ROAD (top) and ROAR (bottom) between pixel-based and super-pixel-based methods. \textbf{Right}: $L_2$-norm difference between super-pixel-based SimpleGrad / SmoothGrad and standard SimpleGrad.}
    \label{fig:fidelity}
\end{figure*}

To explore how the size of super-pixels affects the fidelity and stability, we draw the tradeoff curve between SSIM and $\mu$Fidelity in Fig.~\ref{fig:fidelity}. The experiment is conducted on ImageNet with Quickshift as super-pixel algorithms. We gradually increase the max-distance from 1 to 8. Similar patterns can be observed for all methods, where using small super-pixels for partition can bring large stability gain with only minor drops in terms of fidelity (or even with improvements for both metrics). Too large super-pixels can damage both metrics. For this experiment, dividing the original images into 1000$\sim$2000 groups is a good choice.

\subsection{Interpretability} For applications such as phenomena understanding and knowledge discovery, interpretability is a more important property of saliency maps compared with fidelity. The saliency maps should highlight the important features related to the classification tasks. One way to measure this is to train the same network with the salient features masked and to see if the mask leads to significant performance degeneration. Specifically, we evaluate the interpretability of the pixel-based methods and super-pixel-based methods through Remove and Retrain (ROAR)~\cite{hooker2018evaluating} and Remove and Debias (ROAD)~\cite{rong2022consistent}. ROAR masks top-k pixels with the highest saliency scores (MoRF) and calculates test accuracy on the masked images. ROAD follows a similar process but substitutes retraining with a debiasing operation for faster evaluation. A major and more rapid decrease in test accuracy indicates the masked features are more task-related and the saliency map is more interpretable. The results are shown in Fig. \ref{fig:fidelity}. For both metrics, the drop of super-pixel-based method are faster than their pixel-based counterpart, showing our super-pixel-based method achieves higher interpretability compared to the pixel-based versions.

\section{Discussion and Conclusion}
We note that the strategy followed by our proposed super-
pixel-based method has connections with the SmoothGrad
and GradCAM methods.

\noindent \textbf{Relation to SmoothGrad:} Our partitioning strategy reduces noise variance by averaging importance scores within each pixel partition. This is similar to SmoothGrad, which averages importance scores among multiple noisy versions of the sample using additive Gaussian noise. We compared the $L_2$-norm difference between standard and super-pixel-based SimpleGrad maps as the super-pixel size increased. The curve showed a similar effect on faithfulness as the curve of SmoothGrad with increasing Gaussian noise std. The key difference is that SmoothGrad assumes bounded additive noise does not considerably change the interpretation map, while the super-pixel-based method assumes that pixels within a super-pixel share the same semantic role. These two assumptions are orthogonal and complementary, enhancing the visual quality and stability of the saliency map.

\noindent \textbf{Connection to GradCAM:}
\citet{woerl2023initialization} conduct comprehensive experiments to show GradCAM is less susceptible to initialization noise. Compared with other gradient-based methods, GradCAM generates saliency maps in the semantic layer, which are with lower resolution due to pooling. The lower resolution could be the reason for the stability performance. The super-pixel-based method directly applies partitioning to the input pixels, which align with the texture features. And the comparison with ScoreCAM, which is a variant of GradCAM, also shows the superiority of our method.

\noindent \textbf{Recap on our contribution:}
We proposed a novel super-pixel-based partitioning strategy to explain neural network predictions. Our method improves the quality and stability of saliency maps, as demonstrated by our theoretical evidence and numerical results on various image datasets. Our future work entails applying our method to non-image datasets and addressing robustness against input perturbations and domain shifts in the future.

\section*{Acknowledgements}

The work of Farzan Farnia is partially supported by a grant
from the Research Grants Council of the Hong Kong Special Administrative Region, China, Project 14209920, and
is partially supported by a CUHK Direct Research Grant
with CUHK Project No. 4055164.
\bibliography{egbib}
\end{document}